\pdfoutput=1
\documentclass[letterpaper, 10 pt, conference]{ieeeconf}
\IEEEoverridecommandlockouts
\usepackage{graphicx}
\usepackage{amsmath,amssymb,amsfonts}
\usepackage{booktabs}
\usepackage{float}
\usepackage{dblfloatfix}
\usepackage{cite}

\usepackage[colorlinks=true, linkcolor=blue, citecolor=blue, urlcolor=blue, breaklinks=true]{hyperref}
\hypersetup{pdftitle={SenseSwarm: Swarm Exploration via Neural Structure Estimation with Shared World-Anticipation for Rapid Mapping},
  pdfauthor={Haohua Que, Haojia Gao, Mingkai Liu, Hoiian Au, Qian Zhang, Jiayue Xie, Yule Cai, Bokui Chen, Xueqian Wang, Fei Qiao}}
\graphicspath{{figs/}}

\title{\LARGE
\textbf{SenseSwarm}: \textbf{S}warm \textbf{E}xploration via \textbf{N}eural \textbf{S}tructure \textbf{E}stimation\\
with \textbf{S}hared \textbf{W}orld-\textbf{A}nticipation for \textbf{R}apid \textbf{M}apping}

\author{Haohua Que$^{1,\dagger}$, Haojia Gao$^{2,\dagger}$, Mingkai Liu$^{3}$, Hoiian Au$^{1}$, Qian Zhang$^{1}$,\\ Jiayue Xie$^{1}$, Yule Cai$^{1}$, Bokui Chen$^{4}$, Xueqian Wang$^{2}$, and Fei Qiao$^{1,*}$%
\thanks{$^{\dagger}$These authors contributed equally. $^{*}$Corresponding author: Fei Qiao.}%
\thanks{$^{1}$Department of Electronic Engineering, Tsinghua University, Beijing, China. $^{2}$Tsinghua Shenzhen International Graduate School (SIGS), Tsinghua University, Shenzhen, China. $^{3}$School of Software and Microelectronics, Peking University, Beijing, China. $^{4}$Peng Cheng Laboratory, Shenzhen, China. {\tt\small qiaofei@tsinghua.edu.cn}}%
}

\newcommand{\projsite}{https://que121.github.io/senseswarm-site/}

\begin{document}

\maketitle
\thispagestyle{empty}
\pagestyle{empty}

\begin{abstract}
A robot swarm covers a building faster with more robots, but the map is only as complete as the space the robots have physically visited. We target the amount of an unknown building a decentralized swarm knows at each instant, ahead of physical coverage. SenseSwarm serves a lightweight learned map predictor on the swarm's shared occupancy map and folds the predicted structure back into a stigmergic guidance field. Its 709K-parameter predictor is fine-tuned on the swarm's own multi-front observations, which we show is necessary: a single-robot predictor collapses on such views. We build a benchmark of twelve real building floors from HouseExpo, HKUST, and KTH, up to 24{,}000 square meters, and run 1{,}620 simulations across nine coordination methods, swarm sizes 32 to 512, and three seeds. On physical coverage the swarm is near parity with a greedy frontier planner, as expected. On map reconstruction, every non-predictive baseline needs 1.3 to 1.9 times longer to reach an 80 percent known map, and SenseSwarm holds an equal-time area lead of about 670 square meters, up to 1{,}359. Folding the prediction into the shared field matters: the same predictor applied per robot matches our speed but suffers a 6.9 times higher inter-robot contact rate at 512 robots. The prediction is computed once per shared map, so its cost stayed flat from 32 to 512 robots.
\end{abstract}

\section{Introduction}

Multi-robot exploration of an unknown building is a core capability for search and rescue, inspection, and mapping \cite{cadena2016}. A larger team explores faster, yet the useful output is the map, and the map is only as complete as the space the robots have physically sensed: every unentered room stays blank until a robot walks up to it. We target this ceiling and ask how much of the building a swarm can \emph{know} at each instant, ahead of physical coverage, rather than how fast it can walk every cell.

Prior work optimizes traversal. Frontier methods drive robots to the known-unknown boundary \cite{yamauchi,yamauchi98,burgard2005}, next-best-view planners select informative viewpoints \cite{bircher2016}, decentralized systems such as RACER and TARE coordinate teams to high efficiency \cite{racer,tare}, and benchmarks standardize the metrics \cite{explorebench}. All report how fast the robots cover the space; to our knowledge, none reports how fast a decentralized swarm reconstructs the map when unseen structure is predicted rather than visited.

Map prediction closes exactly this gap, but for single robots: generative networks predict unknown structure and planners act on it \cite{shrestha2019,katyal2019,ramakrishnan2020}, MapEx inpaints the global grid \cite{mapex}, PIPE Planner adds pathwise information gain \cite{pipe}, and SenseMapNet with SenseExpo distill a lightweight local predictor \cite{mapfs,senseexpo}. For multiple robots, 4CNet runs a per-robot diffusion predictor conditioned on teammate trajectories, demonstrated on three robots \cite{4cnet}. No method reports a map-reconstruction speedup for a decentralized swarm. That slot is empty, and it is the one we fill.\footnote{Videos of every run, the benchmark maps, and the construction scripts: \url{\projsite}}

Reaching it requires a predictor that runs at swarm scale, stays accurate on what a swarm actually observes (thin multi-front strips of a shared map, not a single robot's connected blob), and guides the swarm without central control. SenseSwarm answers with one predictor served on the shared map, tiled so the cost tracks the frontier and not the robot count, fine-tuned on the swarm's own observations, and deposited as a stigmergic field that each robot reads only locally.

We evaluate in ARGoS \cite{argos} on twelve real floors \cite{houseexpo,aydemir2012} with 32 to 512 robots, against seven baselines covering every non-predictive decentralized family \cite{smmr,minpos,dsbpf,slug} plus an integration ablation that runs the same predictor per robot, standing in for the 4CNet mechanism at swarm scale.
Our contributions are threefold.
\begin{enumerate}
\item \textbf{The first swarm-scale exploration benchmark on real super-large buildings, to our knowledge.} Twelve real floors up to 24{,}000 square meters, in a 1{,}620-run matrix \cite{explorebench}.
\item \textbf{Predictive stigmergy.} One predictor serves the shared map and becomes a stigmergic field read only locally, so the cost tracks the frontier, not the robot count. It must match the swarm's observation distribution: trained for a single robot, the same architecture drops from 0.85 to 0.34 IoU.
\item \textbf{Reconstruction is where prediction pays.} Every non-predictive baseline needs 1.3 to 1.9 times longer to reach an 80 percent known map, and folding the prediction per robot costs 6.9 times the contacts at 512 robots.
\end{enumerate}

\section{Related Work}
\label{sec:related}

\textbf{Multi-robot exploration.} Frontier exploration sends robots to the known-unknown boundary \cite{yamauchi,yamauchi98}, coordinated variants assign frontiers across a team \cite{burgard2005,simmons2000}, information-theoretic planners maximize expected gain \cite{stachniss2005,bircher2016}, and recent decentralized systems reach high efficiency \cite{racer,tare}. All report coverage time, treating the map as the set of visited cells; SenseSwarm keeps a greedy frontier drive and adds a reconstruction objective prior work does not measure.

\textbf{Learned map prediction.} Predicting unseen structure lets a robot act on a map larger than what it has sensed \cite{shrestha2019,katyal2019,ramakrishnan2020}: MapEx inpaints the global grid with an ensemble \cite{mapex}, PIPE Planner adds pathwise gain \cite{pipe}, and SenseMapNet with SenseExpo distill a lightweight local predictor for a single explorer \cite{mapfs,senseexpo}. 4CNet runs a per-robot diffusion predictor conditioned on shared trajectories \cite{4cnet}, accurate but seconds per robot per cycle and demonstrated on three robots. SenseSwarm instead predicts once on the shared map, is fine-tuned on the swarm observation distribution, and reports a reconstruction speedup at swarm scale.

\textbf{Stigmergic swarm coordination.} Robot swarms reach thousand-robot scale in hardware \cite{rubenstein2014}, and social insects coordinate without central control through stigmergy, signals deposited in the shared environment \cite{grasse1959}. Pheromone-inspired robotics uses shared fields for dispersion and coverage \cite{brambilla2013}, recent methods refine the potential shaping \cite{dsbpf} and the stigmergic gradient itself \cite{slug}, and dense swarms are known to congest past an optimal size \cite{soma2023,hunt2019,fontllenas2018}. We run representatives of both recent families as baselines. Unlike prior stigmergic explorers, SenseSwarm injects a learned predictive signal into the field, pulling the swarm toward structure it has not yet sensed.

\section{Problem Setup and Metrics}
\label{sec:setup}

A swarm of $N$ ground robots explores an unknown planar building. Each robot has a limited-range sensor and writes sensed cells into a shared occupancy grid with three states per cell: free, obstacle, and unknown. Let $\mathcal{F}$ be the set of truly free cells. At time $t$, let $S(t)$ be the sensed-free cells and let $P(t)$ be the cells the predictor marks free with confidence. We measure two quantities.

\emph{Physical coverage} is $C(t) = |S(t) \cap \mathcal{F}| / |\mathcal{F}|$, the fraction of the building the swarm has actually sensed. \emph{Reconstruction} is $R(t) = |(S(t) \cup P(t)) \cap \mathcal{F}| / |\mathcal{F}|$, the fraction the swarm knows once prediction is folded in. Reconstruction is never below coverage and reaches a target level earlier when the predictor is correct.

We summarize the gain three ways, all from the per-frame curves. The \emph{time to known} is the time for a method's known map to reach 80 percent, where known means $R(t)$ for predictive methods and $C(t)$ for baselines. The \emph{equal-time lead} is the reconstruction advantage in percentage points when the best baseline reaches a given sensed level, and the \emph{area lead} converts it into square meters via the floor's free area. Speed of traversal is the time for $C(t)$ to reach 95 percent; safety is the inter-robot contact rate per robot per 100 meters.

\section{Method}
\label{sec:method}

\begin{figure*}[t]
\centering
\includegraphics[width=\textwidth]{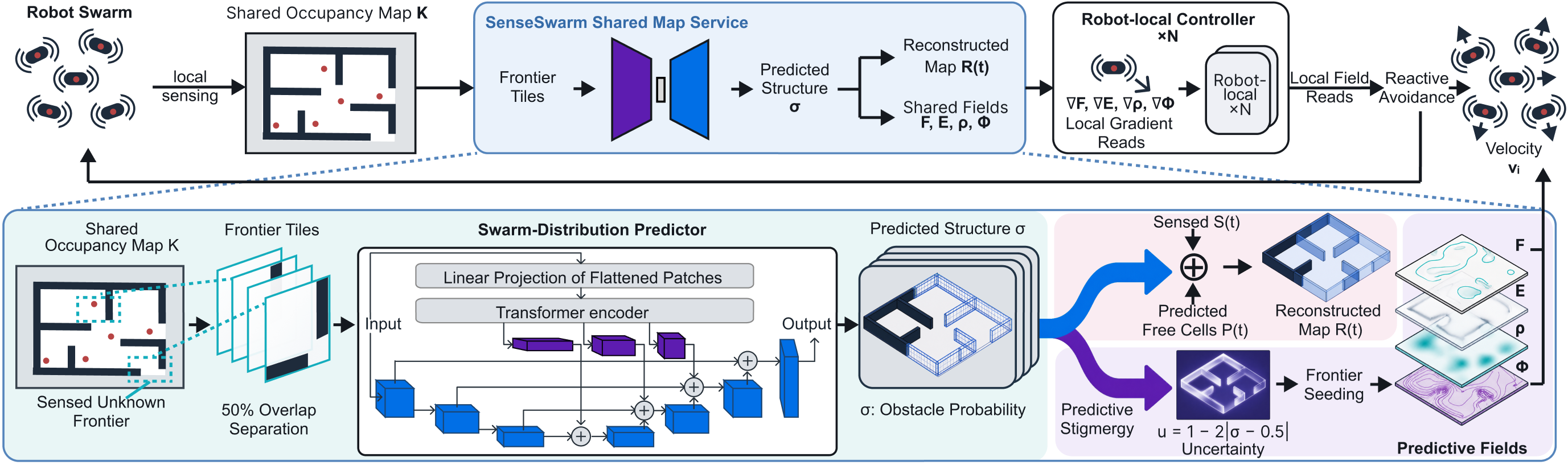}
\caption{The SenseSwarm loop (top) and the shared map service (bottom). Robots sense locally into the shared occupancy map $K$. The service tiles the frontier of $K$ with 50 percent overlap, runs the swarm-distribution predictor once per shared map, and publishes two products of the predicted structure $\sigma$: the predicted free cells $P(t)$ folded with the sensed map $S(t)$ into the reconstruction $R(t)$, and the uncertainty $u = 1 - 2|\sigma - 0.5|$ that seeds the predictive field $\Phi$ at the frontier. Each robot reads only the four field gradients at its own position and steers by Eq.~\eqref{eq:guidance} under reactive avoidance, so the per-robot cost is independent of the swarm size.}
\label{fig:overview}
\end{figure*}

SenseSwarm has three parts: a shared stigmergic map, a swarm-distribution predictor served on that map, and a guidance rule that turns local field gradients into motion. Figure~\ref{fig:overview} shows the full loop; Algorithm 1 summarizes one tick. The predicted structure completes rooms ahead of the sensed frontier.

\begin{figure}[t]
\centering
\begin{minipage}{0.98\linewidth}
\hrule height 0.9pt \vspace{2pt}
{\small \textbf{Algorithm 1} One SenseSwarm tick} \vspace{2pt}
\hrule height 0.5pt \vspace{3pt}
{\small
\begin{tabbing}
1:\ \ \= \textbf{Shared map service:} \\
2: \> integrate all newly sensed cells into occupancy grid $K$ \\
3: \> update frontier-distance field $F$ (BFS from frontiers) \\
4: \> decay and deposit explore trace $E$, density trace $\rho$ \\
5: \> \textbf{if} $t \bmod T_{\mathrm{pred}} = 0$ \textbf{then} \\
6: \> \ \ \= tile the frontier neighborhood of $K$; batch-predict $\sigma$ \\
7: \> \> $u \gets 1 - 2\lvert\sigma - 0.5\rvert$ on predicted unknown cells \\
8: \> \> seed each frontier cell $c$ with $-\lambda \sqrt{\textstyle\sum_{\mathrm{disc}(c)} u}$ \\
9: \> \> propagate seeds into predicted field $\Phi$ (Dijkstra) \\
10: \> \> mark confident predicted obstacles ($\sigma > 0.7$) blocked \\
11: \> \textbf{end if} \\
12: \> \textbf{Each robot $i$ (local reads only):} \\
13: \> $\mathbf{v}_i \gets w_f \nabla F + w_e \nabla E + w_d \nabla \rho + \frac{w_p}{1 + \kappa \rho} \nabla \Phi$ at own cell \\
14: \> blend $\mathbf{v}_i$ with reactive avoidance; move; sense; write $K$
\end{tabbing}}
\vspace{1pt} \hrule height 0.9pt
\end{minipage}
\end{figure} 

\subsection{Shared map and stigmergic fields}
Robots write sensed cells into a shared occupancy grid $K_t : \mathcal{G} \to \{\mathrm{free}, \mathrm{obst}, \mathrm{unk}\}$ and maintain three fields over it, in the manner of a stigmergic coordinator. The frontier set is the sensed-free boundary of the unknown,
\begin{equation}
\mathcal{B}_t = \{ c : K_t(c) = \mathrm{free},\ \exists\, c' \in \mathcal{N}(c) : K_t(c') = \mathrm{unk} \},
\end{equation}
and the frontier-distance field is the geodesic distance to it through sensed-free space,
\begin{equation}
F_t(c) = \min_{b \in \mathcal{B}_t} d_{\mathrm{geo}}(c, b),
\label{eq:frontier}
\end{equation}
so its negative gradient pulls robots toward the nearest reachable frontier. The explore trace $E$ and the robot-density trace $\rho$ are decaying deposits written at visited and occupied cells,
\begin{equation}
X_{t+1}(c) = \gamma_X X_t(c) + \textstyle\sum_{i} \mathbf{1}[c_i(t) = c], \qquad X \in \{E, \rho\},
\label{eq:traces}
\end{equation}
with a slow decay for $E$ (a memory of coverage) and a fast one for $\rho$ (a memory of crowding). Each robot reads only the local gradients of these fields at its own cell, so coordination is decentralized and mediated through the shared environment, not through robot-to-robot messaging.

\subsection{Swarm-distribution prediction, served on the shared map}
The predictor is the 709K-parameter SenseMapNet \cite{mapfs}, a lightweight encoder-decoder $f_\theta$ that maps a local occupancy window $W_c$ of the shared map, centered at $c$ and one-hot encoded over $\{\mathrm{free}, \mathrm{obst}, \mathrm{unk}\}$, to a per-cell obstacle probability
\begin{equation}
\sigma(c') = f_\theta(W_c)(c') \in [0, 1], \qquad c' \in W_c .
\label{eq:pred}
\end{equation}
We serve it once per shared map rather than once per robot. The frontier neighborhood is tiled with overlapping windows, and only tiles that contain a frontier and enough observed context are predicted. This makes the prediction cost scale with the frontier length, which is bounded by the building, and not with $N$. A single inference service answers batched tile requests, so any tested swarm size issues one coherent prediction per update. Measured at the same map state (40 percent known on the largest KTH composite), the full tiled prediction takes 0.39 to 0.40 seconds per update whether the swarm has 32 or 512 robots, a cost ratio of 0.97 while the robot count grows 16 times (401, 395, 402, 391, and 391 ms at $N = 32$ to $512$). A per-robot scheme at the reported seconds per robot per cycle of a diffusion predictor \cite{4cnet} would grow linearly and exceed the exploration timescale itself at swarm scale. Thresholding $\sigma$ also yields the predicted free cells $P(t)$ that Section~\ref{sec:setup} folds into the reconstruction $R(t)$: the same served prediction that guides the swarm below is the map we score.

The predictor must match what the swarm observes. A single robot builds a connected explored blob, while a swarm produces thin multi-front strips that are sparse early and patchy. On a held-out split of exactly such swarm observations, the original single-robot predictor scores 0.34 obstacle IoU in the unknown region. We therefore fine-tune the same architecture on the swarm's own shared-map observations, extracted by replaying stigmergic runs and paired with ground-truth structure, using an obstacle-weighted objective plus a soft skeleton term \cite{cldice} that keeps thin walls connected. This raises the IoU to 0.85. The skeleton term is not cosmetic: without it the same recipe reaches 0.73 IoU and shatters predicted walls into 3.18 times the ground-truth number of connected components, which a planner would read as doorways that do not exist. The fine-tuning recipe is standard; the point is the distribution. A swarm cannot borrow a single-robot map predictor, and to our knowledge this failure mode has not been reported before.

\subsection{Predictive stigmergy: the prediction becomes a field}
The core of SenseSwarm is how the prediction enters the coordination. It is not a plan and not a goal assignment. It is deposited into the shared environment as a fourth stigmergic field, and three design choices make that field useful at swarm scale.

\emph{Uncertainty seeding.} Each sensed frontier cell is seeded with an information value derived from predictive uncertainty rather than predicted area. With per-cell uncertainty $u(c) = 1 - 2\lvert\sigma(c) - 0.5\rvert$, which peaks where the predictor is unsure, each frontier cell $b$ is seeded with
\begin{equation}
s(b) = -\lambda \sqrt{\textstyle\sum_{c \in D_r(b)} u(c)},
\label{eq:seed}
\end{equation}
the summed uncertainty over a disk $D_r$ of radius $r = 5$ m, square-rooted to damp overconfident hallucinations \cite{shrestha2019}: frontiers opening genuinely unresolved structure become deep wells; rooms the predictor has confidently completed do not pull. The seeds propagate into the predicted-frontier field by a generalized geodesic distance over sensed and predicted free space,
\begin{equation}
\Phi_t(c) = \min_{b \in \mathcal{B}_t} \left[ s(b) + d_{\mathrm{geo}}(c, b) \right],
\label{eq:phi}
\end{equation}
so descending $\Phi$ is information-greedy rather than distance-greedy frontier selection: a robot trades travel distance against predicted information gain. We ablate this choice against naive predicted-area seeding in Section~\ref{sec:results}.

\emph{Contention discounting and guidance.} Robot $i$ at cell $c_i$ combines the four fields through normalized negative gradients,
\begin{equation}
\mathbf{v}_i = w_f \widehat{\nabla} F + w_e \widehat{\nabla} E + w_d \widehat{\nabla} \rho + \frac{w_p}{1 + \kappa\, \rho(c_i)}\, \widehat{\nabla} \Phi ,
\label{eq:guidance}
\end{equation}
where $\widehat{\nabla} X$ denotes the unit descent direction of field $X$ sampled at $c_i$, and the predicted pull is divided by the local crowding $1 + \kappa \rho$, so a predicted room that already attracts a crowd stops pulling. The dispersion itself, however, comes from the stigmergic substrate: the explore and density terms of Eq.~\eqref{eq:guidance}, which the per-robot integration ablation lacks entirely. Our ablations in Section~\ref{sec:results} show the substrate is the first-order term and the discount a refinement.

\emph{Local reads only.} The field is written into the shared map like any other stigmergic trace, and each robot reads one local gradient. No robot knows which robot, or which prediction, produced the signal. Coordination stays decentralized, and confident predicted obstacles additionally block dead-end corridors before any robot enters them.

\section{Experimental Setup}
\label{sec:exp}

\begin{figure}[t]
\centering
\includegraphics[width=\linewidth]{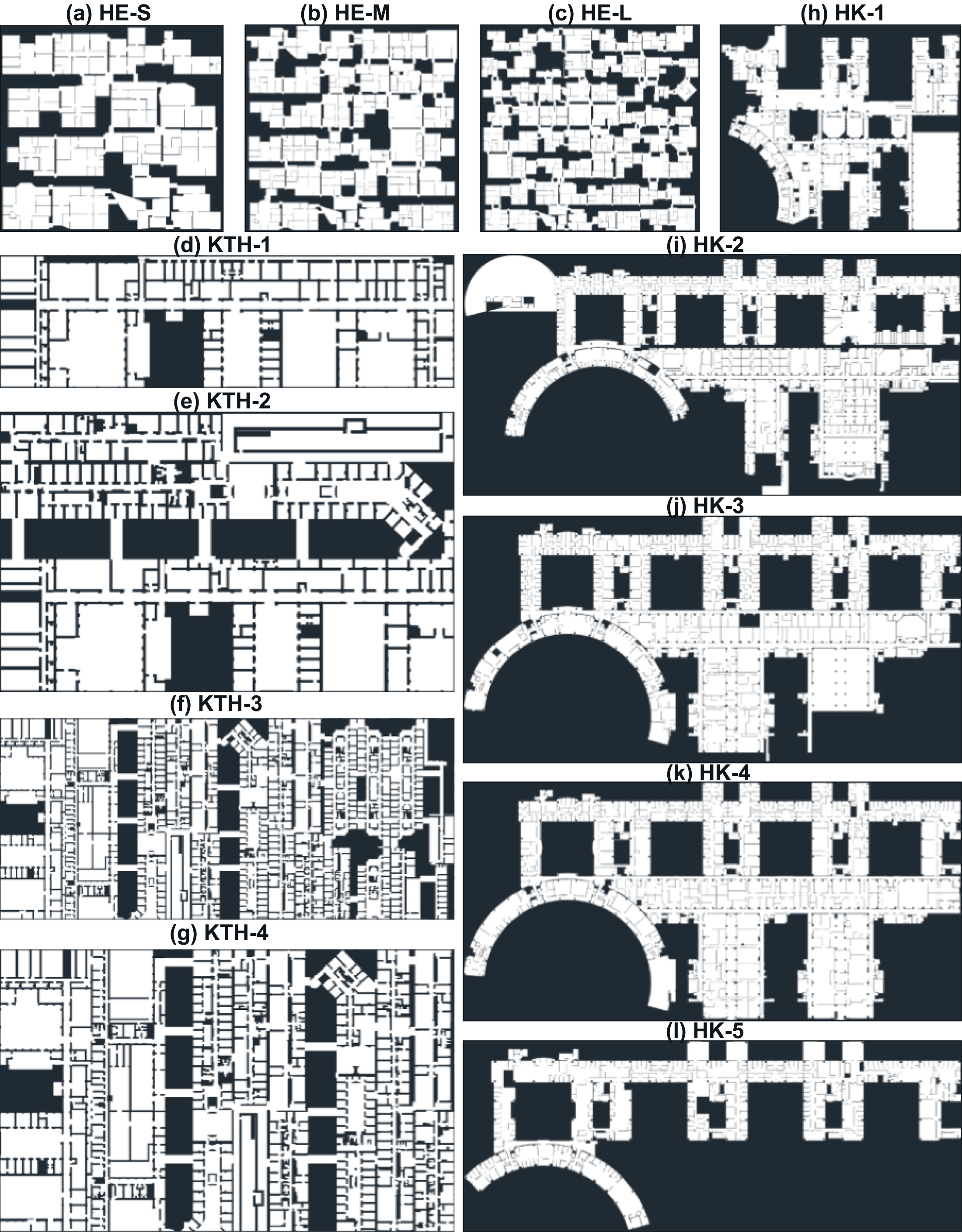}
\caption{The twelve benchmark floors, individually scaled, walls ink and free space white: three HouseExpo apartments (a to c, 2{,}439 to 9{,}574 m$^2$ free), four KTH corridor composites of real stories (d to g, 3{,}752 to 14{,}870 m$^2$), and five HKUST campus floors from the SLABIM building information model (h to l, 10{,}376 to 24{,}015 m$^2$). Per-floor areas are in Table~\ref{tab:main}; full-resolution maps are on the project page.}
\label{fig:benchmark}
\end{figure}

\textbf{Benchmark.} Figure~\ref{fig:benchmark} shows all twelve floors: three HouseExpo apartments \cite{houseexpo} (2{,}400 to 9{,}600 square meters of free space), five HKUST campus floors from the SLABIM building information model (10{,}000 to 24{,}000), and four KTH corridor super-buildings composed from the KTH floorplan dataset \cite{aydemir2012} (3{,}800 to 14{,}900). Both large families follow deterministic construction rules. HKUST floors are rasterized per BIM category, with each door footprint dilated by 1.2 meters to re-open the wall at exactly the real doorway, so rooms connect only where the building's own doors connect them. KTH composites chain uniform-height stories from diverse buildings along one spine, fusing their corridors into a single artery, keeping every original wall, and joining the story bands with 3 meter shafts every 24 meters; the largest stacks twelve stories into a 231 by 102 meter building. All floors are rasterized to a 0.1 meter grid, and every metric is computed over the largest connected free component. The maps, the construction scripts, and videos of every run are on the project page.

\textbf{Methods.} We compare nine coordination methods under an identical reactive avoidance layer. Seven are non-predictive: nearest-frontier \cite{yamauchi98}, MMPF-style potential-field dispersion \cite{smmr}, MinPos frontier allocation \cite{minpos}, a DMPF-Explore-style field that scales frontier attraction down with local crowding \cite{dsbpf}, a Rappel-style dual-layer stigmergic gradient \cite{slug}, a handed-lane variant, and a pure stigmergic coordinator that is exactly SenseSwarm without prediction. The eighth is SenseSwarm. The ninth is not a competitor but our integration ablation: the same predictor applied per robot, standing in for the 4CNet mechanism at swarm scale. MapEx \cite{mapex}, PIPE Planner \cite{pipe}, and 4CNet \cite{4cnet} are single-robot or small-team systems in other simulators and are positioned but not re-run at swarm scale.

\textbf{Protocol.} The full matrix is twelve scenes by nine methods by five swarm sizes (32, 64, 128, 256, 512) by three seeds, for 1{,}620 simulations, all in ARGoS \cite{argos}. Each run dumps the shared map at a fixed interval, and the predictor is applied offline to every frame of every run to produce the per-frame coverage and reconstruction curves from which all metrics are computed. This offline evaluation is identical across methods, so the reconstruction comparison is fair regardless of which method's exploration produced the frames.

\begin{figure*}[t]
\centering
\includegraphics[width=\textwidth]{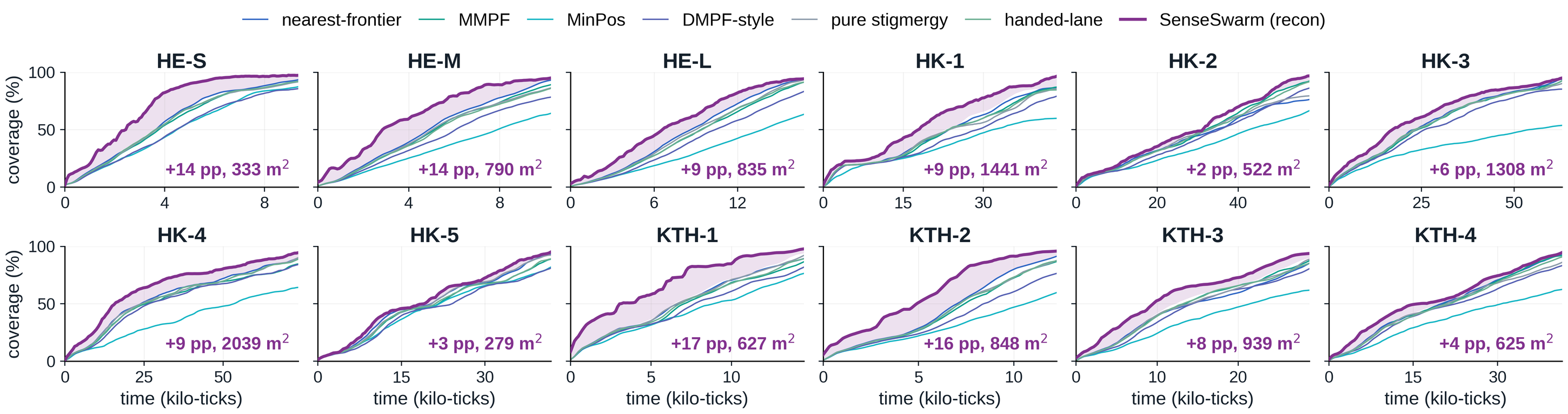}
\vspace{-20pt}
\caption{Coverage versus time on all twelve floors, median over three seeds at 256 robots: thin lines are the six completing baselines' sensed coverage (the settling gradient caps and is omitted), the thick purple line is SenseSwarm's reconstruction, and the shaded band above the best baseline is the building known but not yet visited, annotated as the time-averaged lead in points and square meters.}
\label{fig:curves}
\end{figure*}

\begin{figure*}[t]
\centering
\includegraphics[width=\textwidth]{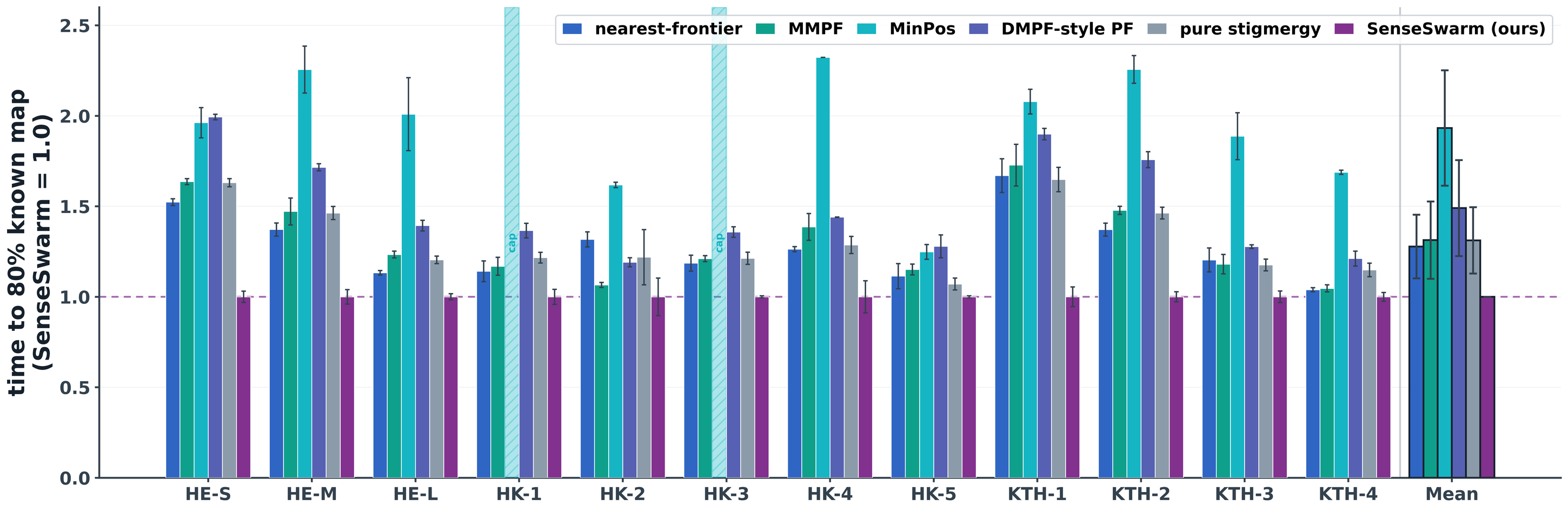}
\caption{Time to an 80 percent known map per floor, normalized so SenseSwarm is 1.0, at 256 robots. Bars are the mean over three seeds and error bars the seed standard deviation; the Mean group averages the twelve floors. Hatched bars marked cap never reach 80 percent. Every non-predictive baseline is above the purple dashed line on nearly every floor: nearest-frontier and MMPF need about 1.3 times our time, the DMPF-style field 1.5, and MinPos 1.9. Pure stigmergy is SenseSwarm without prediction, so its 1.31 bar is the ablated cost of removing the predictor. The per-robot integration ablation is excluded here; it ties our speed and Fig.~\ref{fig:scaling}c shows its safety cost.}
\label{fig:efficiency}
\end{figure*}
\begin{figure*}[t]
\centering
\includegraphics[width=\textwidth]{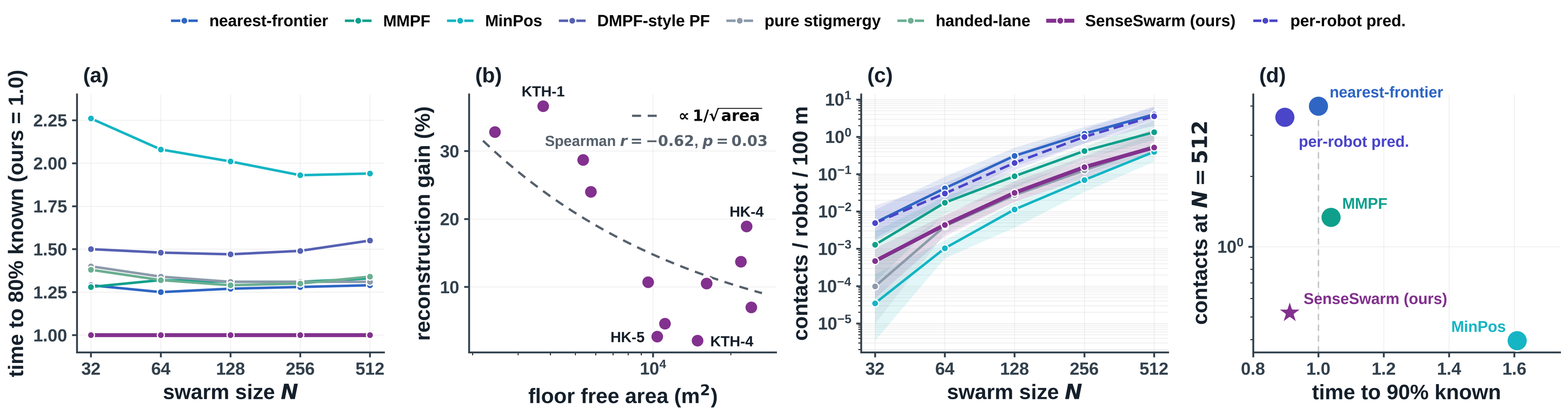}
\vspace{-20pt}
\caption{Scaling and safety, medians over the twelve floors. (a) Normalized time to an 80 percent known map versus swarm size: every baseline holds a flat margin above SenseSwarm from 32 to 512 robots. (b) Per-floor reconstruction gain versus floor area with a $1/\sqrt{\mathrm{area}}$ guide (Spearman $r=-0.62$): the gain shrinks dimensionally with area, and structure modulates the trend (HK-4 versus KTH-4). (c) Contact rate versus swarm size, log-log with interquartile bands, clipped one decade below the median where the lower quartile is zero: per-robot prediction (dashed) congests like uncoordinated nearest-frontier; SenseSwarm stays an order of magnitude lower. (d) Speed versus safety at the harder 90 percent known threshold (a robustness check beyond the 80 percent used elsewhere): time to a 90 percent known map against the contact rate at 512 robots; SenseSwarm alone occupies the fast-and-safe corner.}
\label{fig:scaling}
\end{figure*}

\section{Results}
\label{sec:results}

\subsection{Reconstruction leads physical coverage}
Figure~\ref{fig:efficiency} summarizes the headline result.
Figure~\ref{fig:curves} shows coverage versus time on all twelve floors: the reconstruction leads the best baseline everywhere, and the shaded gap is the building known without having been visited. At 256 robots every non-predictive baseline needs 1.28 to 1.93 times our time to an 80 percent known map. MinPos misses 80 percent on two campus floors, and the settling gradient misses it everywhere: at 512 robots it is blocked 93 percent of its steps against 20 to 35 percent for the rest (Table~\ref{tab:byN}), so it is safe by standing still. Against the per-run best of all seven baselines (Table~\ref{tab:main}), an oracle composite, SenseSwarm still reaches 80 percent known 16 percent sooner on average, up to 37 percent, holding an equal-time lead of 670 square meters and up to 1{,}359; the two units track different building properties (on hkust\_2f a 7 percent speedup still means 1{,}076 square meters), so we report both.

\begin{table}[t]
\centering
\caption{Gain per floor versus the per-run best baseline: time-to-80\%-known reduction and equal-time lead.}
\label{tab:main}
\begin{tabular*}{\linewidth}{@{\extracolsep{\fill}}lrrrr}
\toprule
Floor & Free (m$^2$) & Faster & Lead (pp) & Lead (m$^2$) \\
\midrule
kth\_single & 3{,}752 & $+37\%$ & $+13$ & $+487$ \\
he\_small & 2{,}439 & $+33\%$ & $+13$ & $+320$ \\
kth\_double & 5{,}359 & $+29\%$ & $+13$ & $+685$ \\
he\_medium & 5{,}739 & $+24\%$ & $+10$ & $+557$ \\
hkust\_4f & 23{,}028 & $+19\%$ & $+6$ & $+1{,}359$ \\
hkust\_3f & 21{,}869 & $+14\%$ & $+5$ & $+1{,}055$ \\
he\_large & 9{,}574 & $+11\%$ & $+7$ & $+622$ \\
hkust\_1f & 16{,}106 & $+11\%$ & $+7$ & $+1{,}059$ \\
hkust\_2f & 24{,}015 & $+7\%$ & $+5$ & $+1{,}076$ \\
kth\_triple & 11{,}112 & $+5\%$ & $+3$ & $+344$ \\
hkust\_5f & 10{,}376 & $+3\%$ & $+2$ & $+241$ \\
kth\_quad & 14{,}870 & $+2\%$ & $+2$ & $+233$ \\
\midrule
Mean & & $+16\%$ & $+7$ & $+670$ \\
\bottomrule
\end{tabular*}
\vspace{-9pt}
\end{table}

\begin{table}[t]
\centering
\caption{Time to an 80\% known map per floor at 256 robots (ours = 1.0; cap = never). NF = nearest-frontier, Stig = pure stigmergy, Lane = handed-lane, Set = settling. Bottom: Wilcoxon against each baseline.}
\label{tab:full}
\resizebox{\linewidth}{!}{%
\begin{tabular}{lrrrrrrrr}
\toprule
Floor & NF & MMPF & MinPos & DMPF & Stig & Lane & Set & Ours \\
\midrule
HE-S & 1.52 & 1.64 & 1.96 & 1.99 & 1.63 & 1.64 & cap & 1.00 \\
HE-M & 1.37 & 1.47 & 2.26 & 1.72 & 1.46 & 1.49 & cap & 1.00 \\
HE-L & 1.13 & 1.23 & 2.01 & 1.39 & 1.20 & 1.21 & cap & 1.00 \\
HK-1 & 1.14 & 1.17 & cap & 1.37 & 1.22 & 1.20 & cap & 1.00 \\
HK-2 & 1.32 & 1.07 & 1.62 & 1.19 & 1.22 & 1.09 & cap & 1.00 \\
HK-3 & 1.19 & 1.21 & cap & 1.36 & 1.21 & 1.21 & cap & 1.00 \\
HK-4 & 1.26 & 1.39 & 2.32 & 1.44 & 1.29 & 1.29 & cap & 1.00 \\
HK-5 & 1.11 & 1.15 & 1.25 & 1.28 & 1.07 & 1.13 & cap & 1.00 \\
KTH-1 & 1.67 & 1.73 & 2.08 & 1.90 & 1.65 & 1.69 & cap & 1.00 \\
KTH-2 & 1.37 & 1.48 & 2.26 & 1.76 & 1.46 & 1.47 & cap & 1.00 \\
KTH-3 & 1.20 & 1.18 & 1.89 & 1.28 & 1.18 & 1.18 & cap & 1.00 \\
KTH-4 & 1.04 & 1.05 & 1.69 & 1.21 & 1.15 & 1.07 & cap & 1.00 \\
\midrule
Mean & 1.28 & 1.31 & 1.93 & 1.49 & 1.31 & 1.30 & cap & 1.00 \\
\midrule
Floors ours faster & 12/12 & 12/12 & 10/10 & 12/12 & 12/12 & 12/12 & -- & -- \\
Wilcoxon $p$ & 0.0005 & 0.0005 & 0.0020 & 0.0005 & 0.0005 & 0.0005 & -- & -- \\
\bottomrule
\end{tabular}}
\end{table}

Table~\ref{tab:full} breaks the headline down per floor and per method. SenseSwarm is the fastest method on every floor. The gap is widest on the corridor floors (KTH-1 at 1.67 for nearest-frontier) and narrowest on the two largest composites (KTH-4 at 1.04), consistent with the bounded prediction horizon discussed in Section~\ref{sec:disc}. The ordering is statistically significant (Table~\ref{tab:full}, bottom rows): SenseSwarm beats every baseline on every floor it completes (Wilcoxon signed-rank over the twelve floors, $p \le 0.002$ against each).

\begin{table}[t]
\centering
\caption{Method by swarm size: time to 80\% known and 95\% sensed (ours = 1.0; cap = never), contacts per robot per 100 m, travel (m/m$^2$), and blocked steps (\%). The per-robot ablation ties our speed by construction and pays in contacts.}
\label{tab:byN}
\scriptsize
\resizebox{\linewidth}{!}{%
\begin{tabular}{lrrrrr}
\toprule
Method & 32 & 64 & 128 & 256 & 512 \\
\midrule
\multicolumn{6}{l}{\emph{Time to 80\% known (norm.)}} \\
nearest-frontier & 1.29 & 1.25 & 1.27 & 1.28 & 1.29 \\
MMPF & 1.28 & 1.32 & 1.31 & 1.31 & 1.33 \\
MinPos & 2.26 & 2.08 & 2.01 & 1.93 & 1.94 \\
DMPF-style PF & 1.50 & 1.48 & 1.47 & 1.49 & 1.55 \\
pure stigmergy & 1.40 & 1.34 & 1.31 & 1.31 & 1.31 \\
handed-lane & 1.38 & 1.32 & 1.29 & 1.30 & 1.34 \\
settling gradient & cap & cap & cap & cap & cap \\
per-robot pred. (abl.) & 0.98 & 1.05 & 0.99 & 1.00 & 1.01 \\
SenseSwarm (ours) & 1.00 & 1.00 & 1.00 & 1.00 & 1.00 \\
\midrule
\multicolumn{6}{l}{\emph{Time to 95\% sensed (norm.)}} \\
nearest-frontier & 0.96 & 0.89 & 0.86 & 0.95 & 0.91 \\
MMPF & 1.02 & 0.92 & 0.99 & 1.02 & 1.08 \\
MinPos & 1.40 & 0.92 & 0.81 & 0.81 & 0.90 \\
DMPF-style PF & 0.92 & 0.81 & 1.07 & 1.17 & 1.21 \\
pure stigmergy & 0.99 & 0.92 & 1.06 & 1.04 & 1.06 \\
handed-lane & 1.06 & 1.07 & 0.87 & 1.04 & 1.11 \\
settling gradient & cap & cap & cap & cap & 1.87 \\
per-robot pred. (abl.) & 0.95 & 0.94 & 0.88 & 0.96 & 1.02 \\
SenseSwarm (ours) & 1.00 & 1.00 & 1.00 & 1.00 & 1.00 \\
\midrule
\multicolumn{6}{l}{\emph{Contacts per robot per 100 m}} \\
nearest-frontier & 0.01 & 0.04 & 0.31 & 1.21 & 3.99 \\
MMPF & 0.00 & 0.02 & 0.09 & 0.42 & 1.33 \\
MinPos & 0.00 & 0.00 & 0.01 & 0.07 & 0.40 \\
DMPF-style PF & 0.00 & 0.00 & 0.02 & 0.08 & 0.31 \\
pure stigmergy & 0.00 & 0.00 & 0.03 & 0.13 & 0.56 \\
handed-lane & 0.00 & 0.01 & 0.03 & 0.15 & 0.71 \\
settling gradient & 0.00 & 0.00 & 0.01 & 0.07 & 0.15 \\
per-robot pred. (abl.) & 0.00 & 0.03 & 0.20 & 1.00 & 3.58 \\
SenseSwarm (ours) & 0.00 & 0.00 & 0.03 & 0.15 & 0.52 \\
\midrule
 & Travel$_{256}$ & Travel$_{512}$ & Blocked$_{256}$ & Blocked$_{512}$ & \\
\midrule
nearest-frontier & 0.241 & 0.413 & 17.1 & 20.4 & \\
MMPF & 0.230 & 0.413 & 20.9 & 25.1 & \\
MinPos & 0.339 & 0.585 & 33.9 & 34.9 & \\
DMPF-style PF & 0.257 & 0.461 & 22.9 & 25.8 & \\
pure stigmergy & 0.240 & 0.412 & 20.1 & 23.2 & \\
handed-lane & 0.234 & 0.429 & 21.2 & 24.9 & \\
settling gradient & 0.307 & 0.592 & 94.0 & 93.0 & \\
per-robot pred. (abl.) & 0.227 & 0.396 & 16.9 & 19.5 & \\
SenseSwarm (ours) & \textbf{0.226} & 0.401 & 18.7 & 21.6 & \\
\bottomrule
\end{tabular}}
\vspace{-9pt}
\end{table}
Table~\ref{tab:byN} and Figure~\ref{fig:scaling}a show the same comparison at every swarm size. The margins are stable from 32 to 512 robots: the reconstruction advantage neither relies on a particular density nor erodes at scale, and the pure-stigmergy gap narrows only from 1.40 to 1.31 as larger swarms saturate the floors faster. Figure~\ref{fig:scaling}b shows how the gain scales with the floor instead: it shrinks dimensionally with area, following the bounded-horizon mechanism of Section~\ref{sec:disc}, with structure pushing individual floors off the trend in both directions.

\subsection{Physical coverage is near parity}
On physical coverage SenseSwarm is near the greedy frontier planner: nearest-frontier's median time to 95 percent sensed is 0.86 to 0.96 of ours, and no coordinated baseline separates from parity by more than about 20 percent, except MinPos at 32 robots (Table~\ref{tab:byN}). A greedy frontier drive is already near time-optimal for covering reachable space, so prediction has little room there. SenseSwarm is nonetheless the most motion-efficient method, at a median 0.226 meters of team travel per square meter sensed against 0.241 for nearest-frontier and 0.339 for MinPos, and stalls barely more than uncoordinated frontier (18.7 versus 17.1 percent blocked). The value of prediction is not walking to every cell faster; it is knowing the cell without walking to it.

\subsection{Ablations: integration, field design, and the predictor}
Figure~\ref{fig:scaling}c shows the contact rate against swarm size. The per-robot ablation runs the identical predictor on each robot's own window and is statistically tied with us on reconstruction speed (Wilcoxon $p = 0.42$), and that is the point: it achieves the same speed dangerously. At 512 robots its contact rate is 3.58 per robot per 100 m, 6.9 times ours (0.52), the level of uncoordinated nearest-frontier (3.99). The per-robot variant pulls many robots toward the same predicted room; the density-discounted shared field spreads the same pull across the frontier. Prediction itself costs no safety: pure stigmergy posts 0.56 against our 0.52. The only methods with lower contact rates (MinPos 0.40, DMPF 0.31, settling 0.15) pay by being 1.5 to 1.9 times slower or never reaching 80 percent. They are safe partly by not getting there.

The pure stigmergic coordinator is exactly SenseSwarm without the predicted-frontier field, so the pair isolates the prediction gain: removing prediction slows the time to an 80 percent known map by 31 percent while the contact rate stays flat (0.56 versus 0.52). The prediction buys knowledge, not risk. Two further ablations probe the field design itself, on he\_medium and kth\_single at 256 and 512 robots (Table~\ref{tab:abl}, top). Removing the contention discount, or replacing uncertainty seeding with naive predicted-area seeding, changes the time to an 80 percent known map by at most 7 percent in either direction, within seed noise, and moves the contact rate modestly and inconsistently across the two floors. We report this openly rather than claim each term: the first-order effects are the prediction entering the shared field at all (31 percent above) and the distribution it is trained on (0.34 to 0.85 IoU), and the safety behavior comes from the stigmergic substrate, not from any single refinement.

\begin{table}[t]
\centering
\caption{Component ablations on he\_medium and kth\_single: guidance design (top) and sensing range (bottom).}
\label{tab:abl}
\resizebox{\linewidth}{!}{%
\begin{tabular}{llrrr}
\toprule
Floor & Metric & full & no discount & area seeding \\
\midrule
he\_medium & time (256) & 1.00 & 0.96 & 1.00 \\
he\_medium & time (512) & 1.00 & 1.00 & 1.01 \\
he\_medium & contacts & 0.50 & 0.59 & 0.63 \\
kth\_single & time (256) & 1.00 & 0.97 & 0.97 \\
kth\_single & time (512) & 1.00 & 1.06 & 1.07 \\
kth\_single & contacts & 1.93 & 1.25 & 1.62 \\
\midrule
Floor & Gain vs frontier & $R = 1.0$ m & $R = 1.5$ m & $R = 2.5$ m \\
\midrule
he\_medium & time to 80\% known & $+33\%$ & $+27\%$ & $+24\%$ \\
kth\_single & time to 80\% known & $+37\%$ & $+40\%$ & $+33\%$ \\
\bottomrule
\end{tabular}}
\vspace{-9pt}
\end{table}

The gain also survives sensor degradation (Table~\ref{tab:abl}, bottom): shrinking the sensing radius from 2.5 to 1.0 meters raises the gain over nearest-frontier from 24 to 33 percent on he\_medium, since weaker sensors leave more unsensed and prediction fills exactly that.

\begin{figure}[t]
\centering
\includegraphics[width=\linewidth,height=0.84\textheight,keepaspectratio]{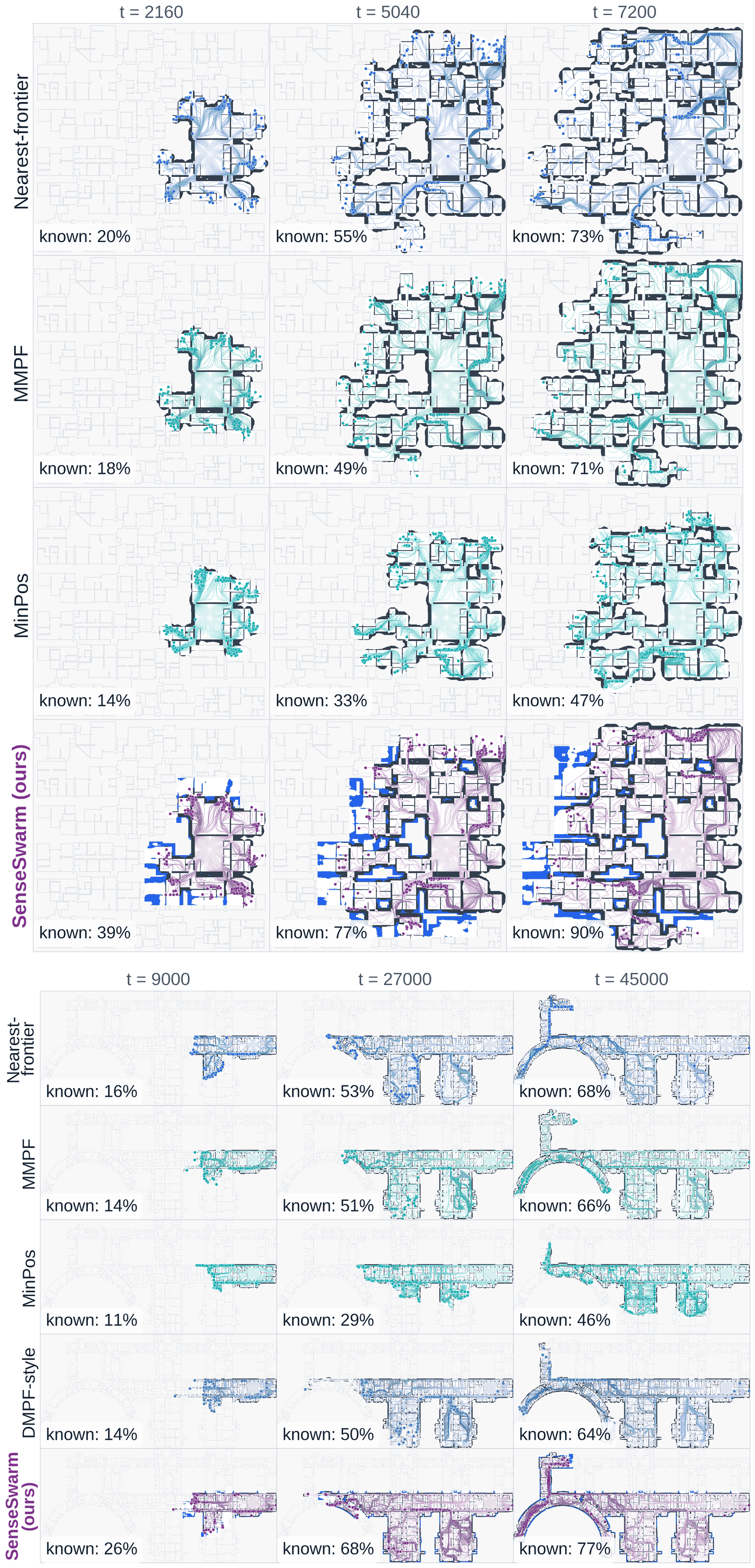}
\caption{Coordination methods at 256 robots: the same apartment floor (top) and campus scale, the 23{,}000 square meter hkust\_4f (bottom). Path colors match the method colors of Figs.~\ref{fig:curves} to \ref{fig:scaling} and darken with time; SenseSwarm is purple. Chips state the known map fraction, which for the non-predictive baselines equals their sensed coverage; predicted structure is blue, and unknown space carries a faint ground-truth outline; nothing faint is visible to any method. Trails link logged positions and omit chords that would cross a wall between samples. Baselines bunch along corridors; SenseSwarm disperses toward predicted rooms and leads nearest-frontier's known map by 9 to 15 points at equal campus ticks.}
\label{fig:methods}
\end{figure}
\begin{figure}[t]
\centering
\includegraphics[width=\linewidth]{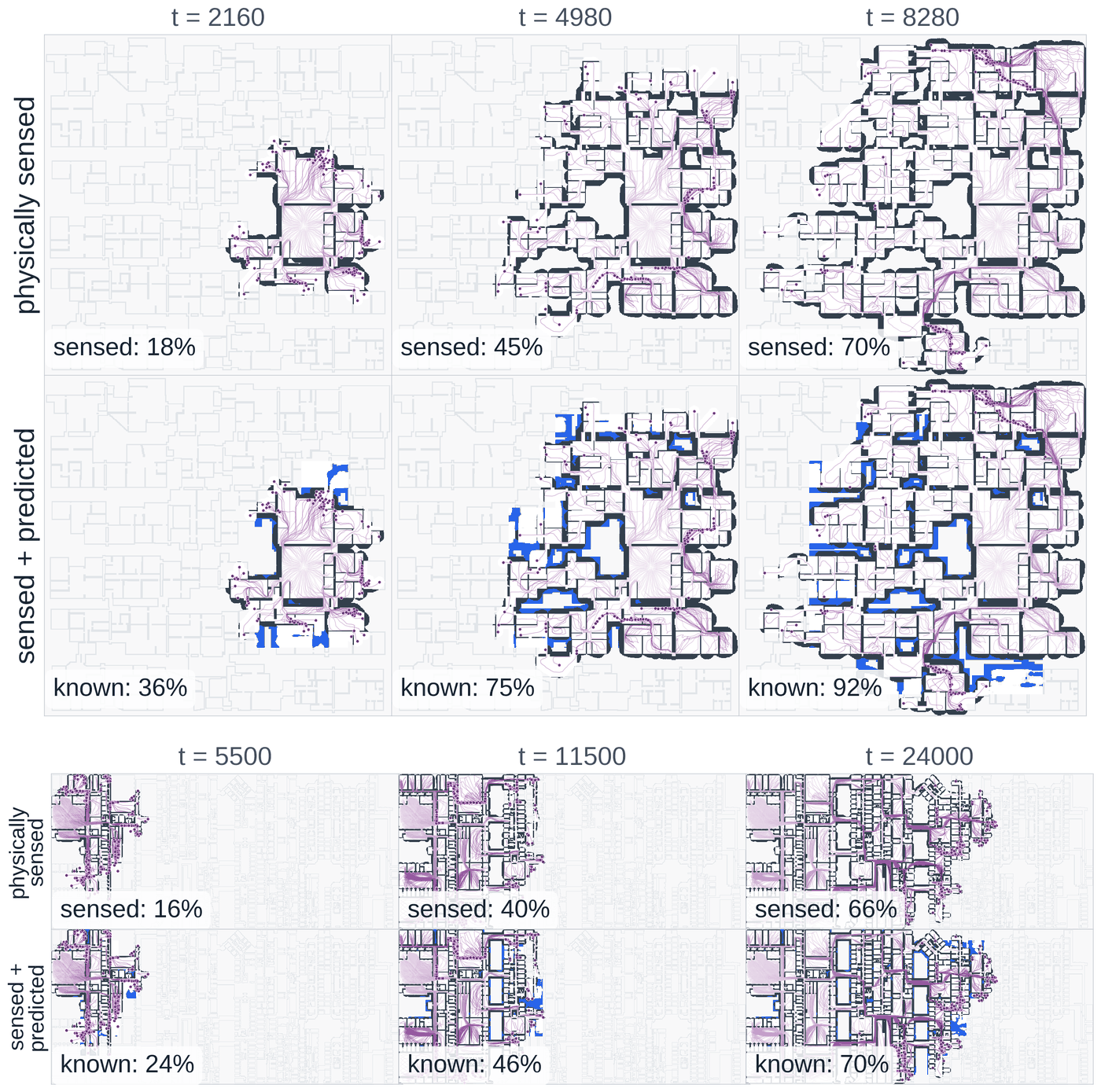}
\caption{Sensed map versus reconstruction over time: a HouseExpo apartment floor at 256 robots (top block) and the twelve-story kth\_quad composite, 231 meters long, at 512 robots (bottom block), the hardest floor for prediction, whose stories connect through a few shafts crossed serially (Sec.~\ref{sec:disc}). SenseSwarm paths are purple and darken with time, predicted structure is blue, and unknown space carries the faint ground-truth outline. Rooms appear in the reconstruction several panels before any robot enters them, and even the shaft-serialized composite stays 4 to 8 points ahead.}
\label{fig:recon_he}
\end{figure}

Finally, the predictor generalizes to buildings it has never seen (Table~\ref{tab:predictor}): retrained with hkust\_3f excluded and re-evaluated offline on the same frames, the leave-one-building-out predictor keeps 10.8 of the paired predictor's 14.4 percent gain. Three quarters of the gain survives on a floor the predictor never saw, so it comes from learned building regularities, not memorized floors.

\begin{table}[t]
\centering
\caption{Predictor ablations: obstacle IoU on the held-out swarm split (top); leave-one-building-out gain on hkust\_3f (bottom).}
\label{tab:predictor}
\begin{tabular*}{\linewidth}{@{\extracolsep{\fill}}lr}
\toprule
Observation distribution & obstacle IoU \\
\midrule
single-robot original & 0.34 \\
swarm fine-tune, obstacle-weighted BCE & 0.73 \\
swarm fine-tune, {}+{} soft clDice (ours) & 0.85 \\
\midrule
Generalization (held-out hkust\_3f) & gain \\
\midrule
paired predictor, floor in training & $+14.4\%$ \\
leave-one-building-out, floor never seen & $+10.8\%$ \\
\bottomrule
\end{tabular*}
\vspace{-9pt}
\end{table}

The predictor must also be trained on the right distribution: the 0.34 versus 0.85 IoU gap in Table~\ref{tab:predictor} is a 2.5 times difference from the observation distribution alone. The predictor is not more accurate in general. It is matched to what the swarm sees, and that match is what makes the reconstruction lead possible.

\subsection{Qualitative behavior}
Figure~\ref{fig:methods} contrasts the main coordination methods mid-run on an apartment floor and repeats the comparison at campus scale, where the parity-and-lead pattern is visible directly: equal sensed coverage at every shared tick, with the known map ahead by 9 to 15 points. Figure~\ref{fig:recon_he} shows the sensed-versus-reconstructed pair over time on an apartment floor and scales the same view to the twelve-story composite at 512 robots. The baselines' paths bunch along corridors and re-traverse covered rooms, while SenseSwarm's paths fan out early toward rooms the predictor has completed but no robot has entered. Figure~\ref{fig:scaling}d compresses the speed and safety results into a single view: SenseSwarm alone occupies the fast-and-safe corner.

\section{Discussion and Limitations}
\label{sec:disc}

The reconstruction gain shrinks with floor area (Spearman $r = -0.62$, $p = 0.03$), and the mechanism is dimensional, not a predictor failure. The tiled prediction extends knowledge a bounded band beyond the frontier, so the absolute lead grows with frontier length, roughly $\sqrt{\mathrm{area}}$, and it does grow, from 320 to 1{,}359 square meters. The baseline's time to 80 percent instead grows with the area itself: on large floors the bottleneck shifts from observation to travel, and prediction saves observation. Structure modulates the trend both ways, from 19 percent on the 23{,}000 square meter hkust\_4f to 2 percent on kth\_quad, whose stories connect through a few shafts the frontier must pass serially. Breaking the bounded horizon with hierarchical prediction \cite{mapex} is future work.

Our evaluation is in simulation, on planar indoor floors that are static within a run, with an idealized shared map: updates propagate instantly, and communication range, latency, and loss are not modeled, in line with prior swarm-exploration work. A physical deployment needs a shared-map service or virtual-stigmergy replication \cite{pinciroli2016buzz} and a common localization reference; our per-cell updates are monotonic, so merges are conflict-free and the incremental bandwidth is small. Extending to 3D structure, dynamic scenes, realistic radio models, and cm-scale hardware is future work.

\section{Conclusion}
\label{sec:conc}

We presented SenseSwarm, a prediction-enhanced decentralized swarm explorer that serves a lightweight swarm-distribution predictor once on the shared occupancy map and folds the predicted structure into a stigmergic field. On twelve real floors and 1{,}620 simulations, prediction buys little on traversal, where a greedy planner is already near optimal, but a large gain on reconstruction: every non-predictive baseline needs 1.3 to 1.9 times longer to reach an 80 percent known map, removing prediction costs 31 percent, and the contact rate stays 6.9 times below per-robot prediction at 512 robots. The result reframes what prediction is for in swarm exploration: not to walk the building faster, but to know it before you walk it.

\section*{Acknowledgment}
The authors thank the Sense Lab for computing resources.

\end{document}